\documentclass[conference, letterpaper]{IEEEtran}
\ifCLASSINFOpdf
\else
\fi

\usepackage{booktabs}
\usepackage{tikz}
\usetikzlibrary{arrows,decorations.pathmorphing,backgrounds,positioning,fit,petri,shapes,automata}

\usepackage{subcaption}

\usepackage[cmex10]{amsmath}
\usepackage{amsfonts}
\usepackage{color}
\usepackage{fancyhdr}

\renewcommand{\thispagestyle}[2]{}

\fancypagestyle{plain}{
        \fancyhead{}
        \fancyhead[C]{first page center header}
        \fancyfoot{}
        \fancyfoot[C]{first page center footer}
}
\newtheorem{definition}{Definition}

\begin{document}

%
% paper title
% can use linebreaks \\ within to get better formatting as desired
\title{Event Abstraction for Process Mining using Supervised Learning Techniques}

% author names and affiliations
% use a multiple column layout for up to three different
% affiliations
%\author{\IEEEauthorblockN{Anonymized}
%\IEEEauthorblockA{}}
\author{\IEEEauthorblockN{Niek Tax}
\IEEEauthorblockA{Eindhoven University of Technology\\ Department of Mathematics and Computer Science\\ P.O. Box 513, 5600MB Eindhoven\\ The Netherlands\\Email: n.tax@tue.nl}\\[0.25cm]
\IEEEauthorblockN{Natalia Sidorova}
\IEEEauthorblockA{Eindhoven University of Technology\\ Department of Mathematics and Computer Science\\ P.O. Box 513, 5600MB Eindhoven\\ The Netherlands\\Email: n.sidorova@tue.nl}
\and
\IEEEauthorblockN{Reinder Haakma}
\IEEEauthorblockA{Philips Research\\ Prof. Holstlaan 4\\ 5665 AA Eindhoven\\ The Netherlands\\Email: reinder.haakma@philips.com}\\[0.25cm]
\IEEEauthorblockN{Wil M. P. van der Aalst}
\IEEEauthorblockA{Eindhoven University of Technology\\ Department of Mathematics and Computer Science\\ P.O. Box 513, 5600MB Eindhoven\\ The Netherlands\\Email: w.m.p.v.d.aalst@tue.nl}}

% make the title area
\maketitle

\begin{abstract}
Process mining techniques focus on extracting insight in processes from event logs. In many cases, events recorded in the event log are too fine-grained, causing process discovery algorithms to discover incomprehensible process models or process models that are not representative of the event log. We show that when process discovery algorithms are only able to discover an unrepresentative process model from a low-level event log, structure in the process can in some cases still be discovered by first abstracting the event log to a higher level of granularity. This gives rise to the challenge to bridge the gap between an original low-level event log and a desired high-level perspective on this log, such that a more structured or more comprehensible process model can be discovered. We show that supervised learning can be leveraged for the event abstraction task when annotations with high-level interpretations of the low-level events are available for a subset of the sequences (i.e., traces). We present a method to generate feature vector representations of events based on XES extensions, and describe an approach to abstract events in an event log with Condition Random Fields using these event features. Furthermore, we propose a sequence-focused metric to evaluate supervised event abstraction results that fits closely to the tasks of process discovery and conformance checking. We conclude this paper by demonstrating the usefulness of supervised event abstraction for obtaining more structured and/or more comprehensible process models using both real life event data and synthetic event data.
\end{abstract}
% IEEEtran.cls defaults to using nonbold math in the Abstract.
% This preserves the distinction between vectors and scalars. However,
% if the conference you are submitting to favors bold math in the abstract,
% then you can use LaTeX's standard command \boldmath at the very start
% of the abstract to achieve this. Many IEEE journals/conferences frown on
% math in the abstract anyway.

% no keywords


% For peer review papers, you can put extra information on the cover
% page as needed:
% \ifCLASSOPTIONpeerreview
% \begin{center} \bfseries EDICS Category: 3-BBND \end{center}
% \fi
%
% For peerreview papers, this IEEEtran command inserts a page break and
% creates the second title. It will be ignored for other modes.
\IEEEpeerreviewmaketitle

\section{Introduction}
Process mining is a fast growing discipline that combines knowledge and techniques from computational intelligence, data mining, process modeling and process analysis \cite{Aalst2011}. Process mining focuses on the analysis of event logs, which consists of sequences of real-life events observed from process executions, originating e.g. from logs from ERP systems. An important subfield of process mining is process discovery, which is concerned with the task of finding a process model that is representative of the behavior seen in an event log. Many different process discovery algorithms exist (\cite{Aalst2004,Gunther2007,Werf2008,Weijters2011,Leemans2013}), and many different types of process models can be discovered by process discovery methods, including BPMN models, Petri nets, process trees, and statecharts. 

\begin{figure}
	\centering
	\includegraphics[width=0.48\textwidth]{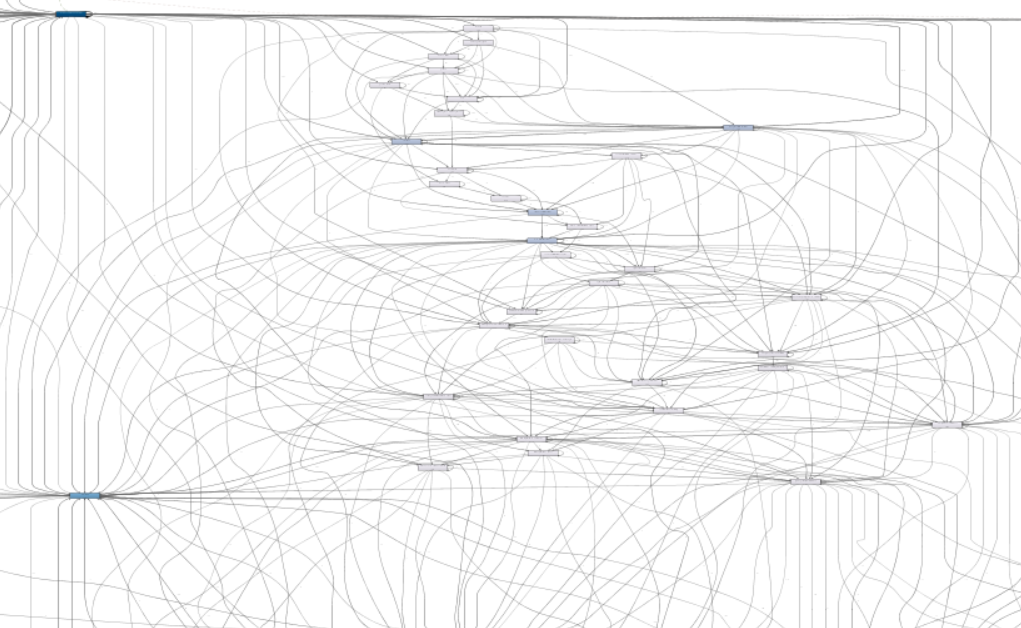}
	\caption{An excerpt of a "spaghetti"-like process model.}
	\label{fig:spaghetti}
\end{figure}

\begin{figure}
	\centering
	\includegraphics[width=0.25\textwidth]{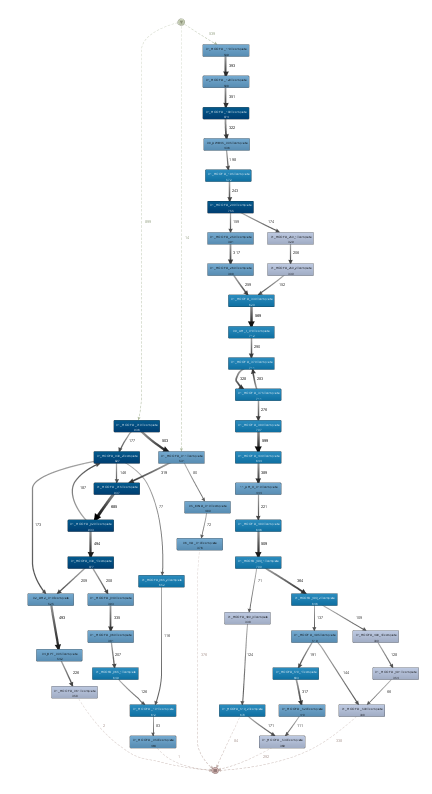}
	\caption{A structured, or "lasagna"-like, process model.}
	\label{fig:lasagna}
\end{figure}

As event logs are often not generated specifically for the application of process mining, events granularity of the event log at hand might be too low level. It is vital for successful application of process discovery techniques to have event logs at an appropriate level of abstraction. Process discovery techniques when the input event log is too low level might result in process model with one or more undesired properties. First of all, the resulting process model might be "spaghetti"-like, as shown in Figure \ref{fig:spaghetti}, containing of an uninterpretable mess of nodes and connections. The aim of process discovery is to discover a structured, "lasagna"-like, process model as shown in Figure \ref{fig:lasagna}, which is much more interpretable than a "spaghetti"-like model. Secondly, the activities in the process model might have too specific, non-meaningful, names. Third, as we show in section \ref{sec:motivating_example}, process discovery algorithms are sometimes not able to discover a process model that represents the low-level event log well, while being able to discover to discover a representative process model from a corresponding high-level event log. The problems mentioned illustrate the need for a method to abstract too low-level event logs into higher level event logs. 

Several methods have been explored within the process mining field that address the challenge of abstracting low-level events to higher level events (\cite{Bose2009,Gunther2010,Dongen2010}). Existing event abstraction methods rely on unsupervised learning techniques to abstract low-level into high-level events by clustering together groups of low-level events into one high-level event. However, using unsupervised learning introduces two new problems. First, it is unclear how to label high-level events that are obtained by clustering low-level events. Current techniques require the user / process analyst to provide high-level event labels themselves based on domain knowledge, or generate long labels by concatenating the labels of all low-level events incorporated in the cluster. However, long concatenated labels quickly become unreadable for larger clusters, and it is far from trivial for a user to come up with sensible labels manually. In addition, unsupervised learning approaches for event abstraction give no guidance with respect to the desired level of abstraction. Many existing event abstraction methods contain one or more parameters to control the degree in which events are clustered into higher level events. Finding the right level of abstraction providing meaningful results is often a matter of trial-and-error.

In some cases, training data with high-level target labels of low-level events are available, or can be obtained, for a subset of the traces. In many settings, obtaining high-level annotations for all traces in an event log is infeasible or too expensive. Learning a supervised learning model on the set of traces where high-level target labels are available, and applying that model to other low-level traces where no high-level labels are available, allows us to build a high-level interpretation of a low-level event log, which can then be used as input for process mining techniques.

In this paper we describe a method for supervised event abstraction that enables process discovery from too fine-grained event logs. This method can be applied to any event log where higher level training labels of low level events are available for a subset of the traces in the event log. We start by giving an overview of related work from the activity recognition field in Section \ref{sec:related}. In Section \ref{sec:preliminaries} we introduce basic concepts and definitions used throughout the rest of the paper. Section \ref{sec:motivating_example} explains the problem of not being able to mine representative process models from low-level data in more detail. In Section \ref{sec:features} we describe a method to automatically retrieve a feature vector representation of an event that can be used with supervised learning techniques, making use of certain aspects of the XES standard definition for event logs \cite{Gunther2014}. In the same section we describe a supervised learning method to map low-level events into target high-level events. Sections \ref{sec:case_1} and \ref{sec:case_2} respectively show the added value of the described supervised event abstraction method for process mining on a real life event log from a smart home environment and on a synthetic log from a digital photocopier respectively. Section \ref{sec:conclusion} concludes the paper.

\section{Related Work}
\label{sec:related}
Supervised event abstraction is an unexplored problem in process mining. A related field is activity recognition within the field of ubiquitous computing. Activity recognition focuses on the task of detecting human activity from either passive sensors \cite{Kasteren2008,Tapia2004}, wearable sensors \cite{Bao2004,Kwapisz2011}, or cameras \cite{Poppe2010}. %Discrete events from passive or wearable sensors can be considered to be low-level events, while the human activities can be regarded as high-level events.
Activity recognition methods generally work on discrete time windows over the time series of sensor values and aim to map each time window onto the correct type of human activity, e.g. \emph{eating} or \emph{sleeping}. Activity recognition methods can be classified into probabilistic approaches \cite{Kasteren2008,Tapia2004,Bao2004,Kwapisz2011} and approaches based on ontology reasoning \cite{Chen2009,Riboni2011}. The strength of probabilistic system based approaches compared to methods based on ontology reasoning is their ability to handle noise, uncertainty and incomplete in sensor data \cite{Chen2009}.

Tapia \cite{Tapia2004} was the first to explore supervised learning methods to infer human activity from passive sensors, using a naive Bayes classifier. More recently, probabilistic graphical models started to play an important role in the activity recognition field \cite{Kasteren2008,Kasteren2007}. Van Kasteren et al. \cite{Kasteren2008} explored the use Conditional Random Fields \cite{Lafferty2001} and Hidden Markov Models \cite{Rabiner1986}. Van Kasteren and Kr{\"o}se \cite{Kasteren2007} applied Bayesian Networks \cite{Friedman1997} on the activity recognition task. Kim et al. \cite{Kim2010} found Hidden Markov Models to be incapable of capturing long-range or transitive dependencies between observations, which results in difficulties recognizing multiple interacting activities (concurrent or interwoven). Conditional Random Fields do not posses these limitations.

The main differences between existing work in activity recognition and the approach presented in this paper are the input data on which they can be applied and the generality of the approach. Activity recognition techniques consider the input data to be a multidimensional time series of the sensor values over time based on which time windows are mapped onto human activities. An appropriate time window size is determined based on domain knowledge of the data set. In supervised event abstraction we aim for a generic method that works for all XES event logs in general. A time window based approach contrasts with our aim for generality, as no single time window size will be appropriate for all event logs. Furthermore, the durations of the events within a single event log might differ drastically (e.g. one event might take seconds, while another event takes months), in which case time window based approaches will either miss short events in case of larger time windows or resort to very large numbers of time windows resulting in very long computational time. Therefore, we map each individual low-level event to a high-level event and do not use time windows. In a smart home environment context with passive sensors, each change in a binary sensor value can be considered to be a low-level event.

\section{Preliminaries}
\label{sec:preliminaries}
In this section we introduce basic concepts used throughout the paper.

We use the usual sequence definition, and denote a sequence by listing its elements, e.g. we write $\langle a_1,a_2,\dots,a_{n} \rangle$ for a (finite) sequence $s:\{1,\dots,n\}\to S$ of elements from some alphabet $S$, where $s(i)=a_i$ for any $i \in \{1,\dots,n\}$.
\subsection{XES Event Logs}
\begin{figure}
	\centering
	\includegraphics[width=0.95\linewidth]{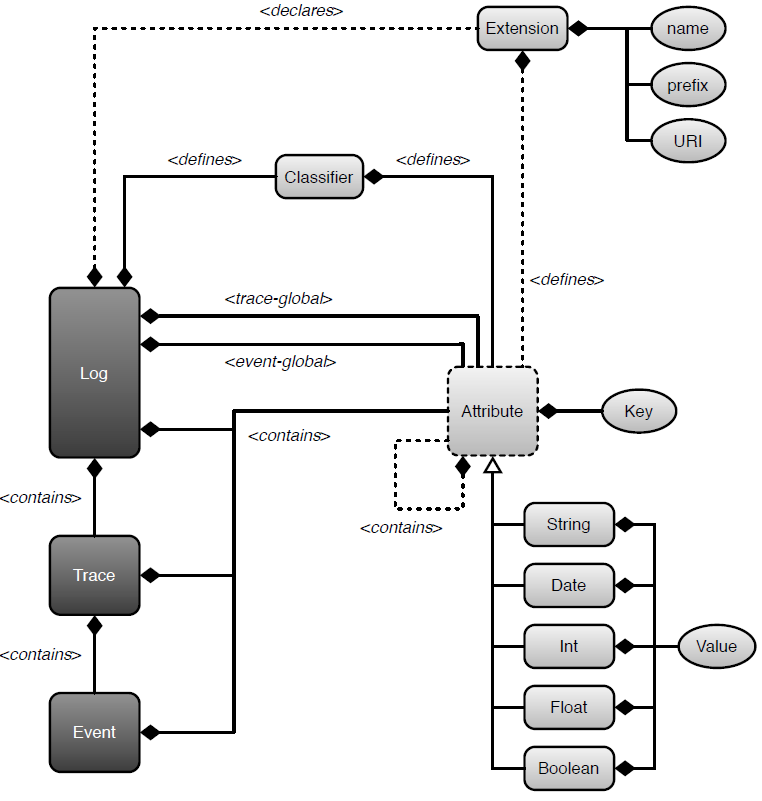}
	\caption{XES event log meta-model, as defined in \cite{Gunther2014}.}
	\label{fig:XES_metamodel}
\end{figure}

We use the XES standard definition of event logs, an overview of which is shown in Figure \ref{fig:XES_metamodel}. XES defines an event \emph{log} as a set of \emph{traces}, which in itself is a sequence of \emph{event}s. The log, traces and events can all contain one or more \emph{attribute}s, which consist of a \emph{key} and a \emph{value} of a certain type. Event or trace attributes may be \emph{global}, which indicates that the attribute needs to be defined for each event or trace respectively. A log contains one or more \emph{classifier}s, which can be seen as labeling functions on the events of a log, defined on global event attributes. \emph{Extension}s define a set of attributes on log, trace, or event level, in such a way that the semantics of these attributes are clearly defined. One can view XES extensions as a specification of attributes that events, traces, or event logs themselves frequently contain. XES defines the following standard extensions:
\begin{description}[\IEEEsetlabelwidth{Organizational}]
	\item[Concept] {Specifies the generally understood name of the event/trace/log (attribute 'Concept:name').}
	\item[Lifecycle] {Specifies the lifecycle phase (attribute 'Lifecycle:transition') that the event represents in a transactional model of their generating activity. The \emph{Lifecycle} extension also specifies a standard transactional model for activities.}
	\item[Organizational]{Specifies three attributes for events, which identify the actor having caused the event (attribute 'Organizational:resource'), his role in the organization (attribute 'Organizational:role'), and the group or department within the organization where he is located (attribute 'Organizational:group').}
	\item[Time]{Specifies the date and time at which an event occurred (attribute 'Time:timestamp').}
	\item[Semantic]{Allows definition of an activity meta-model that specifies higher-level aggregate views on events (attribute 'Semantic:modelReference').}
\end{description}
We introduce a special attribute of type \emph{String} with key \emph{label}, which represents a high-level version of the generally understood name of an event. The \emph{concept} name of a event is then considered to be a low-level name of an event. The \emph{Semantic} extension closely resembles the \emph{label} attribute, however, by specifying relations between low-level and high-level events in a meta-model, the \emph{Semantic} extension assumes that all instances of a low-level event type belong to the same high-level event type. The \emph{label} attribute specifies the high-level label for each event individually, allowing for example one low-level event of low-level type \emph{Dishes \& cups cabinet} to be of high-level type \emph{Taking medicine}, and another low-level event of the same type to be of high-level type \emph{Eating}. Note that for some traces high-level annotations might be available, in which case its events contain the \emph{label} attribute, while other traces might not be annotated. High-level interpretations of unannotated traces, by inferring the \emph{label} attribute based on information that is present in the annotated traces, allow the use of unannotated traces for process discovery and conformance checking on a high level.

\subsection{Petri nets}
A process modeling notation frequently used as output of process discovery techniques is the Petri net. Petri nets are directed bipartite graphs consisting of transitions and places, connected by arcs. Transitions represent activities, while places represent the status of the system before and after execution of a transition. Labels are assigned to transitions to indicate the type of activity that they represent. A special label $\tau$ is used to represent invisible transitions, which are only used for routing purposes and do not represent any real activity.
\begin{definition}[Labeled Petri net]
	\label{def:lpn}
	A labeled Petri net is a tuple $N=(P,T,F,R,\ell)$ where $P$ is a finite set of places, $T$ is a \emph{finite set} of transitions such that $P \cap T = \emptyset$, and $F \subseteq (P \times T) \cup (T \times P)$ is a set of directed arcs, called the flow relation, $R$ is a finite set of labels representing event types, with $\tau \notin R$ is a label representing an invisible action, and $\ell:T\rightarrow R\cup {\tau}$ is a labeling function that assigns a label to each transition.
\end{definition}

The state of a Petri net is defined w.r.t. the state that a process instance can be in during its execution. A state of a Petri net is captured by the marking of its places with tokens. In a given state, each place is either empty, or it contains a certain number of tokens. A transition is enabled in a given marking if all places with an outgoing arc to this transitions contain at least one token. Once a transition fires (i.e. is executed), a token is removed from all places with outgoing arcs to the firing transition and a token is put to all places with incoming arcs from the firing transition, leading to a new state.  
\begin{definition}[Marking, Enabled transitions, and Firing]
	A marked Petri net is a pair $(N,M)$, where $N=(P,T,F,L,\ell)$ is a labeled Petri net and where $M \in \mathbb{B}(P)$ denotes the marking of $N$. For $n \in (P \cup T)$ we use $\bullet n$ and $n \bullet$ to denote the set of inputs and outputs of n respectively. Let $C(s,e)$ indicate the number of occurrences (count) of element $e$ in multiset $s$. A transition $t\in T$ is enabled in a marking $M$ of net $N$ if $\forall p \in \bullet t : C(M,p)>0$. An enabled transition $t$ may fire, removing one token from each of the input places $\bullet t$ and producing one token for each of the output places $t\bullet$. %We use $M_1 \overset{t}{\rightarrow} M_2$ to denote that a transition $t$ is enabled in state $M_1$ and after firing results in state $M_2$. For a firing sequence $\sigma$, $M_1 \overset{\sigma}{\rightarrow} M_2$ indicates that $M_2$ can be reached from $M_1$ through firing sequence $\sigma$.
\end{definition}

Figure \ref{fig:double_flower} shows three Petri nets, with the circles representing places, the squares representing transitions. The black squares represent invisible transitions, or, $\tau$ transitions. Places annotated with an \textbf{f} belong to the final marking, indicating that the process execution can terminate in this marking.

The topmost Petri net in Figure \ref{fig:double_flower} initially has one token in the place $p1$, indicated by the dot. Firing of silent transition $t1$ takes the token from $p1$ and puts a token in both $p2$ and $p3$, enabling both $t2$ and $t3$. When $t2$ fires, it takes the token from $p2$ and puts a token in $p4$. When $t3$ fires, it takes the token from $p3$ and puts a token in $p5$. After $t2$ and $t3$ have both fired, resulting in a token in both $p4$ and $p5$, $t4$ is enabled. Executing $t4$ takes the token from both $p4$ and $p5$, and puts a token in $p6$. The \textbf{f} indicates that the process execution can stop in the marking consisting of this place. Alternatively, it can fire $t5$, taking the token from $p6$ and placing a token in $p2$ and $p5$, which allows for execution of $MC$ and $W$ to reach the marking consisting of $p6$ again. We refer the interested reader to \cite{Reisig2012} for an extensive review of Petri nets.

\subsection{Conditional Random Field}
\label{sec:crf}
We view the recognition of high-level event labels as a sequence labeling task in which each event is classified as one of the higher-level events from a high-level event alphabet. Conditional Random Fields (CRFs) \cite{Lafferty2001} are a type of probabilistic graphical model which has become popular in the fields of language processing and computer vision for the task of sequence labeling. A Conditional Random Field models the conditional probability distribution of the label sequence given an observation sequence using a log-linear model. We use Linear-chain Conditional Random Fields, a subclass of Conditional Random Fields that has been widely used for sequence labeling tasks, which takes the following form:\\ %that make a first-order Markov assumption among label variables $Y$. More formally, a Conditional Random Field defines the conditional probability of label sequence $Y$ given observation sequence $X$ as

$p(y|x) = \frac{1}{Z(x)}exp(\sum_{t=1}\sum_k\lambda_k f_k(t,y_{t-1},y_t,x))$\\

where $Z(x)$ is the normalization factor, $X=\langle x_1,\dots,x_n\rangle$ is an observation sequence, $Y=\langle y_1,\dots,y_n\rangle$ is the associated label sequence, $f_k$ and $\lambda_k$ respectively are feature functions and their weights. Feature functions, which can be binary or real valued, are defined on the observations and are used to compute label probabilities. In contrast to Hidden Markov Models \cite{Rabiner1986}, feature functions are not assumed to be mutually independent. %The feature weights in Conditional Random Fields can be optimized using gradient based methods, such as the conjugate gradient method or L-BFGS.
%A Conditional Random Field (CRF) defines a conditional probability distribution $p(Y|X)$ of label sequences given input sequences. We assume that the random variable sequences $X$ and $Y$ have the same length, and use event input sequence $x=\langle x_1,\dots, x_n\rangle$ and event output sequence $y=\langle y_1,\dots,y_n\rangle$. A CRF on $(X,Y)$ is specified by a vector $f$ of local features and a corresponding weight vector $\lambda$. Each local feature is either a start
%\section{The Effect of Unstructured High-level Concepts}
%\label{sec:double_flower}

\section{Motivating Example}
\label{sec:motivating_example}
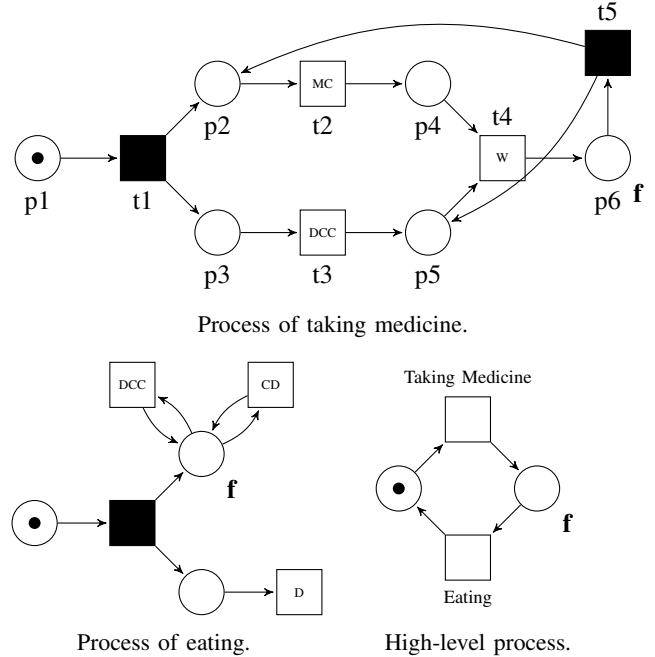
\begin{figure}[t]
	\centering
	\captionsetup[subfloat]{labelformat=empty}
	\begin{subfigure}{0.5\textwidth}
		\centering
		\begin{tikzpicture}
		[node distance=1.4cm,
		on grid,>=stealth',
		bend angle=20,
		auto,
		every place/.style= {minimum size=6mm},
		every transition/.style = {minimum size = 6mm}
		]
		\node [place, tokens = 1] (p2) [label=below:p1]{};
		\node [transition] (2) [fill=black, right = of p2, label=below:t1] {}
		edge [pre] node[auto] {} (p2);
		\node [place] (p3) [above right = of 2, label=below:p2] {}
		edge[pre] node[auto] {} (2);
		\node [place] (p4) [below right = of 2, label=below:p3] {}
		edge[pre] node[auto] {} (2);
		\node [transition] (3) [label=center:\tiny{MC}, label=below:t2, right = of p3] {}
		edge[pre] node[auto] {} (p3);
		\node [transition] (4) [label=center:\tiny{DCC}, label=below:t3, right = of p4] {}
		edge[pre] node[auto] {} (p4);
		\node [place] (p5) [right = of 3, label=below:p4] {}
		edge[pre] node[auto] {} (3);
		\node [place] (p6) [right = of 4, label=below:p5] {}
		edge[pre] node[auto] {} (4);
		\node [transition] (5) [label=center:\tiny{W}, label=above:t4, above right = of p6] {}
		edge[pre] node[auto] {} (p5)
		edge[pre] node[auto] {} (p6);
		\node [place] (p8) [right = of 5, label=below right:\textbf{f}, label=below:p6] {}
		edge[pre] node[auto] {} (5);
		\node [transition] (6) [fill=black,above = of p8, label=above:t5] {}
		edge[pre] node[auto] {} (p8)
		edge[post,bend left] node[auto] {} (p6)
		edge[post,bend right] node[auto] {} (p3);
		%\node [transition] (7) [fill=black,below = of p8] {}
		%edge[pre] node[auto] {} (p8);
		%\node [place] (p9) [left = of 7] {}
		%edge[pre] node[auto] {} (7);
		\end{tikzpicture}
		\caption{Process of taking medicine.}
	\end{subfigure}
	\hspace{0.01\textwidth}
	\begin{subfigure}{0.22\textwidth}
		\centering
		\begin{tikzpicture}
		[node distance=1.3cm,
		on grid,>=stealth',
		bend angle=20,
		auto,
		every place/.style= {minimum size=6mm},
		every transition/.style = {minimum size = 6mm}
		]
		\node [place, tokens = 1] (p1){};
		\node[transition] (0) [right = of p1,fill=black]{}
		edge [pre] node[auto] {} (p1);
		\node [place] (p2)[above right = of 0, label=below right:\textbf{f}]{}
		edge [pre] node[auto] {} (0);
		\node [transition] (2) [label=center:\tiny{CD}, above right = of p2] {}
		edge [pre, bend left] node[auto] {} (p2)
		edge [post, bend right] node[auto] {} (p2);
		\node [transition] (1) [label=center:\tiny{DCC}, above left = of p2] {}
		edge [pre, bend left] node[auto] {} (p2)
		edge [post, bend right] node[auto] {} (p2);
		\node [place] (p3)[below right = of 0]{}
		edge[pre] node[auto] {} (0);
		
		\node [transition] (3) [label=center:\tiny{D}, right = of p3] {}
		edge [pre] node[auto] {} (p3);
		\end{tikzpicture}
		\caption{Process of eating.}
	\end{subfigure}
	\begin{subfigure}{0.22\textwidth}
		\centering
		\begin{tikzpicture}
		[node distance=1.3cm,
		on grid,>=stealth',
		bend angle=20,
		auto,
		every place/.style= {minimum size=6mm},
		every transition/.style = {minimum size = 6mm}
		]
		\node [place, tokens = 1] (p1){};
		edge [pre, bend left] node[auto] {} (p1)
		edge [post, bend right] node[auto] {} (p1);
		\node [transition] (t1) [label=above:\scriptsize{Taking Medicine}, above right = of p1] {}
		edge [pre] node[auto] {} (p1);
		\node [transition] (t2) [label=below:\scriptsize{Eating}, below right = of p1] {}
		edge [post] node[auto] {} (p1);
		\node [place] (p2)[below right = of t1, label=below right:\textbf{f}]{}
		edge [pre] node[auto] {} (t1)
		edge [post] node[auto] {} (t2);
		%\node[transition] (t3) [fill=black, right = of p2] {}
		%edge [pre] node[auto] {} (p2);
		%\node[place] (p3)[right = of t3]{}
		%edge [pre] node[auto] {} (t3);
		\end{tikzpicture}
		\caption{High-level process.}
	\end{subfigure}
	\caption{A high-level process consisting of two unstructured subprocesses that overlap in event types.}
	\label{fig:double_flower}
\end{figure}
Figure \ref{fig:double_flower} shows on a simple example how a process can be structured at a high level while this structure is not discoverable from a low-level log of this process. The bottom right Petri net shows the example process at a high-level. The high-level process model allows for any finite length alternating sequence of \emph{Taking medicine} and \emph{Eating} activities. The \emph{Taking medicine} high-level activity is defined as a subprocess, corresponding to the topmost Petri net, which consists of low-level events \emph{Medicine cabinet (MC)}, \emph{Dishes \& cups cabinet (DCC)}, and \emph{Water (W)}. The \emph{Eating} high-level event is also defined as a subprocess, shown in the bottom left Petri net, which consists of low-level events \emph{Dishes \& cups cabinet (DCC)} and \emph{Cutlery drawer (CD)} that can occur an arbitrary number of times in any order and low-level event \emph{Dishwasher (D)} which occurs exactly once, but at an arbitrary point in the \emph{Eating} process.

\begin{figure*}[t]
	\centering
	\includegraphics[width=0.75\textwidth]{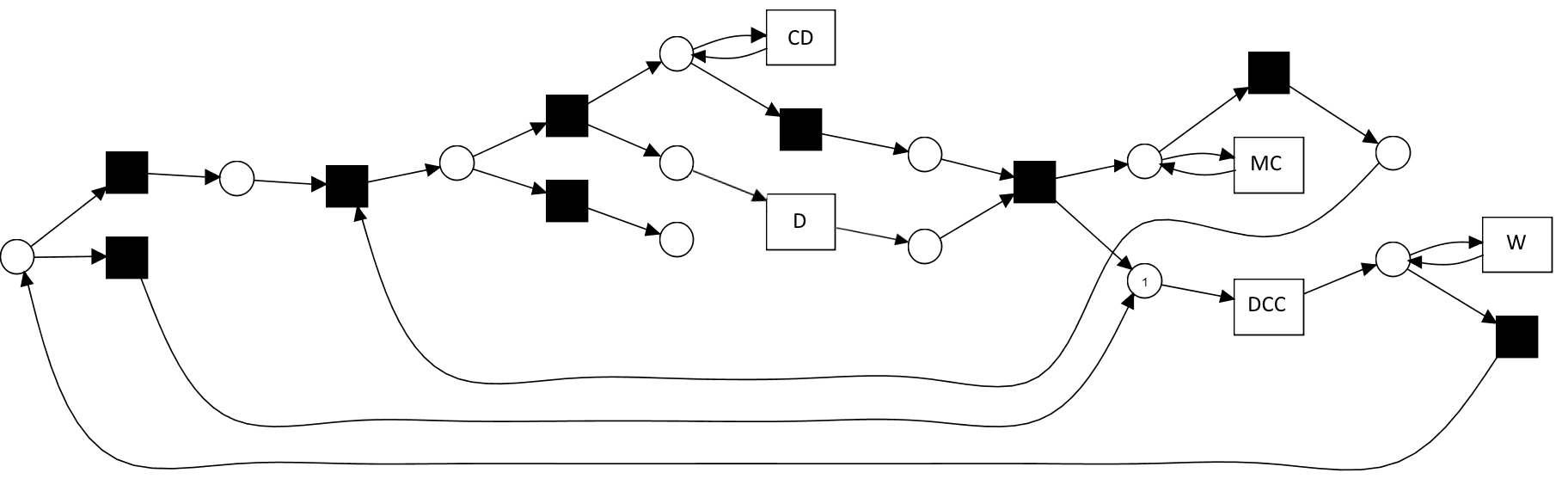}
	\caption{Result of the Inductive Miner on the low-level traces, reduced using Murata reduction rules \cite{Murata1989}.}
	\label{fig:merged_flower}
\end{figure*}

%Consider two random traces generated by this low-level process, $\{\langle A,C,B,C,\\3,1,2,C,C,2,1,1\rangle,\langle C,B,C,1,3,A,C,B,3,2,1,1,B,A\rangle\}$, which at high-level are traces $\{\langle X,Y,X,Y\rangle,\langle X,Y,X,Y,X\rangle\}$.
When we apply the Inductive Miner process discovery algorithm \cite{Leemans2013} to low-level traces generated by the hierarchical process of Figure \ref{fig:double_flower}, we obtain the process model shown in Figure \ref{fig:merged_flower}. The obtained process model allows for almost all possible sequences over the alphabet $\{CD,D,DCC,MC,W\}$, as the only constraint introduced by the model is that \emph{DCC} and \emph{D} are required to be executed starting from the initial marking to end up with the same marking. Firing of all other transitions in the model can be skipped. Behaviorally this model is very close to the so called "flower" model \cite{Aalst2011}, the model that allows for all behavior over its alphabet. The alternating structure between \emph{Taking medicine} and \emph{Eating} that was present in the high-level process in Figure \ref{fig:double_flower} cannot be observed in the process model in Figure \ref{fig:merged_flower}. This is caused by high variance in start and end events of the high-level event subprocesses of \emph{Taking medicine} and \emph{Eating} as well as by the overlap in event types between these two subprocesses. 

When the event log would have consisted of the high-level \emph{Eating} and \emph{Taking medicine} events, process discovery techniques have no problems to discover the alternating structure in the bottom right Petri net of Figure \ref{fig:double_flower}. To discover the high-level alternating structure from a low-level event log it is necessary to first abstract the events in the event log. Through supervised learning techniques the mapping from low-level events to high-level events can be learned from examples, without requiring a hand-made ontology. Similar approaches have been explored in activity recognition in the field of ubiquitous computing, where low-level sensor signals are mapped to high-level activities from a human behavior perspective. The input data in this setting are continuous time series from sensors. Change points in these time series are triggered by low-level activities like \emph{opening/closing the fridge door}, and the annotations of the higher level events (e.g. \emph{cooking}) are often obtained through manual activity diaries. In contrast to unsupervised event abstraction, the annotations in supervised event abstraction provide guidance on how to label higher level events and guidance for the target level of abstraction.

\section{Event Abstraction as a Sequence Labeling Task}
\label{sec:features}
\begin{figure}
	\includegraphics[width=0.5\textwidth]{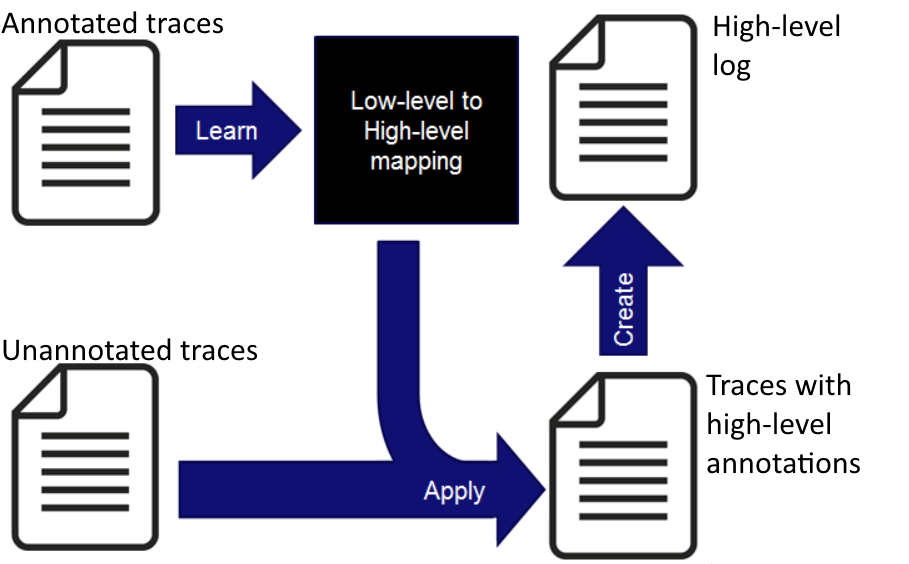}
	\caption{Conceptual overview of Supervised Event Abstraction.}
	\label{fig:overview}
\end{figure}

In this section we describe an approach to supervised abstraction of events based on Conditional Random Fields. Additionally, we describe feature functions on XES event logs in a general way by using XES extensions. Figure \ref{fig:overview} provides a conceptual overview of the supervised event abstraction method. The approach takes two inputs, 1) a set of annotated traces, which are traces where the high-level event that a low-level event belongs to (the \emph{label} attribute of the low-level event) is known for all low-level events in the trace, and 2) a set of unannotated traces, which are traces where the low-level events are not mapped to high-level events. Conditional Random Fields are trained of the annotated traces to create a probabilistic mapping from low-level events to high-level events. This mapping, once obtained, can be applied to the unannotated traces in order to estimate the corresponding high-level event for each low-level event (its \emph{label} attribute). Often sequences of low-level events in the traces with high-level annotations will have the same \emph{label} attribute. We make the working assumption that multiple high-level events are executed in parallel. This enables us to interpret a sequence of identical \emph{label} attribute values as a single instance of a high-level event. To obtain a true high-level log, we \emph{collapse} sequences of events with the same value for the \emph{label} attribute into two events with this value as \emph{concept} name, where the first event has a \emph{lifecycle} 'start' and the second has the \emph{lifecycle} 'complete'. Table \ref{tab:collapse} illustrates this collapsing procedure through an input and output event log.

\begin{table*}
	\centering\scriptsize
	\caption{Left: a trace with predicted high-level annotations (\emph{label}) and, Right: the resulting high-level log after collapsing subsequent identical label values.}
	\begin{subtable}{0.48\linewidth}
		\centering
		\begin{tabular}{llll}
			\toprule
			Case & Time:timestamp & Concept:name & label \\
			\midrule
			1 & 03/11/2015 08:45:23 & Medicine cabinet & Taking medicine\\
			1 & 03/11/2015 08:46:11 & Dishes \& cups cabinet & Taking medicine\\
			1 & 03/11/2015 08:46:45 & Water & Taking medicine\\
			1 & 03/11/2015 08:47:59 & Dishes \& cups cabinet & Eating\\
			1 & 03/11/2015 08:47:89 & Dishwasher & Eating\\
			1 & 03/11/2015 17:10:58 & Dishes \& cups cabinet & Taking medicine\\
			1 & 03/11/2015 17:10:69 & Medicine cabinet & Taking medicine\\
			1 & 03/11/2015 17:11:18 & Water & Taking medicine\\
			\bottomrule
		\end{tabular}
	\end{subtable}
	\begin{subtable}{0.48\linewidth}
		\centering
		\begin{tabular}{llll}
			\toprule
			Case & Time:timestamp & Concept:name & Lifecycle:transition \\
			\midrule
			1 & 03/11/2015 08:45:23 & Taking medicine & Start\\
			1 & 03/11/2015 08:46:45 & Taking medicine & Complete\\
			1 & 03/11/2015 08:47:59 & Eating & Start\\
			1 & 03/11/2015 08:47:89 & Eating & Complete\\
			1 & 03/11/2015 17:10:58 & Taking medicine & Start\\
			1 & 03/11/2015 17:11:18 & Taking medicine & Complete\\
			\bottomrule
		\end{tabular}
	\end{subtable}
	\label{tab:collapse}
\end{table*}

The method described in this section is implemented and available for use as a plugin for the ProM 6 \cite{Verbeek2010} process mining toolkit and is based on the GRMM \cite{Sutton2006} implementation of Conditional Random Fields. 

We now show for each XES extension how it can be translated into useful feature functions for event abstraction. Note that we do not limit ourselves to XES logs that contain all XES extensions; when a log contains a subset of the extensions, a subset of the feature functions will be available for the supervised learning step. This approach leads to a feature space of unknown size, potentially causing problems related to the curse of dimensionality, therefore we use L1-regularized Conditional Random Fields. L1 regularization causes the vector of feature weights to be sparse, meaning that only a small fraction of the features have a non-zero weight and are actually used by the prediction model. Since the L1-norm is non-differentiable, we use OWL-QN \cite{Andrew2007} to optimize the model.

\subsection{From a XES Log to a Feature Space}
\subsubsection{Concept extension}
The low-level labels of the preceding events in a trace can contain useful contextual information for high-level label classification. Based on the n-gram of $n$ last-seen events in a trace, we can calculate the probability that the current event has a label $l$. A multinoulli distribution is estimated for each n-gram of $n$ consecutive events, based on the training data. The Conditional Random Field model requires feature functions with numerical range. A concept extension based feature function with two parameters, $n$ and $l$, is valued with the multinoulli-estimated probability of the current event having high-level label $l$ given the n-gram of the last $n$ low-level event labels.
\subsubsection{Organizational extension}
Similar to the concept extension feature functions, multinoulli distributions can be estimated on the training set for n-grams of \emph{resource}, \emph{role}, or \emph{group} attributes of the last $n$ events. Likewise, an organizational extension based feature function with three parameters, n-gram size $n$, $o\in\{resource,role,group\}$, and label $l$, is valued with the multinoulli-estimated probability of label $l$ given the n-gram of the last $n$ event resources/roles/groups.
\subsubsection{Time extension}
In terms of time, there are several potentially existing patterns. A certain high-level event might for example be concentrated in a certain parts of a day, of a week, or of a month. This concentration can however not be modeled with a single Gaussian distribution, as it might be the case that a high-level event has high probability to occur in the morning or in the evening, but low probability to occur in the afternoon in-between. Therefore we use a Gaussian Mixture Model (GMM) to model the probability of a high-level label $l$ given the timestamp. Bayesian Information Criterion (BIC) \cite{Schwarz1978} is used to determine the number of components of the GMM, which gives the model an incentive to not combine more Gaussians in the mixture than needed. A GMM is estimated on training data, modeling the probabilities of each label based on the time passed since the start of the day, week or month. A time extension based feature function with two parameters, $t\in\{day,week,month,\dots\}$ and label $l$, is valued with the GMM-estimated probability of label $l$ given the $t$ view on the event timestamp.
\subsubsection{Lifecycle extension \& Time extension}
The XES standard \cite{Gunther2014} defines several lifecycle stages of a process. When an event log possesses both the lifecycle extension and the time extension, time differences can be calculated between different stages of the life cycle of a single activity. For a \emph{complete} event for example, one could calculate the time difference with the associated \emph{start} event of the same activity. Finding the associated \emph{start} event becomes nontrivial when multiple instances of the same activity are in parallel, as it is then unknown which \emph{complete} event belongs to which \emph{start} event. We assume consecutive lifecycle steps of activities running in parallel to occur in the same order as the preceding lifecycle step. For example, when we observe two \emph{start} events of an activity of type \emph{A} in a row, followed by two \emph{complete} events of type \emph{A}, we assume the first \emph{complete} to belong to the first \emph{start}, and the second \emph{complete} to belong to the second \emph{start}. 

We estimate a Gaussian Mixture Model (GMM) for each tuple of two lifecycle steps for a certain activity on the time differences between those two lifecycle steps for this activity. A feature based on both the lifecycle and the time extension, with a label parameter $l$ and lifecycle $c$, is valued with the GMM-estimated probability of label $l$ given the time between the current event and lifecycle $c$. Bayesian Information Criterion (BIC) \cite{Schwarz1978} is again used to determine the number of components of the GMM.

\subsection{Evaluating High-level Event Predictions for Process Mining Applications}
Existing approaches in the field of activity recognition take as input time windows where each time window is represented by a feature vector that describes the sensor activity or status during that time window. Hence, evaluation methods in the activity recognition field are window-based, using evaluation metrics like the percentage of correctly classified time slices \cite{Tapia2004,Kasteren2007,Kasteren2008}. There are two reasons to deviate from this evaluation methodology in a process mining setting. First, our method operates on events instead of time windows. Second, the accuracy of the resulting high level sequences is much more important for many process mining techniques (e.g. process discovery, conformance checking) than the accuracy of predicting each individual minute of the day.

We use \emph{Levenshtein similarity} that expresses the degree in which two traces are similar using a metric based on the Levenshtein distance (also known as edit distance) \cite{Levenshtein1966}, which is defined as $Levenshtein\_similarity(a,b)=1-\frac{Levenshtein\_distance(a,b)}{max(|a|,|b|)}$. The division of the Levenshtein distance by $max(|a|,|b|)$, which is the worst case number of edit operations needed to transform any sequence of length $|a|$ into any sequence of length $|b|$, causes the result to be a number between $0$ (completely different traces) and $1$ (identical traces).

\section{Case Study 1: Smart Home Environment}
\label{sec:case_1}

\begin{figure*}
	\centering
	\begin{subfigure}{0.92\textwidth}
		\includegraphics[width=0.92\textwidth]{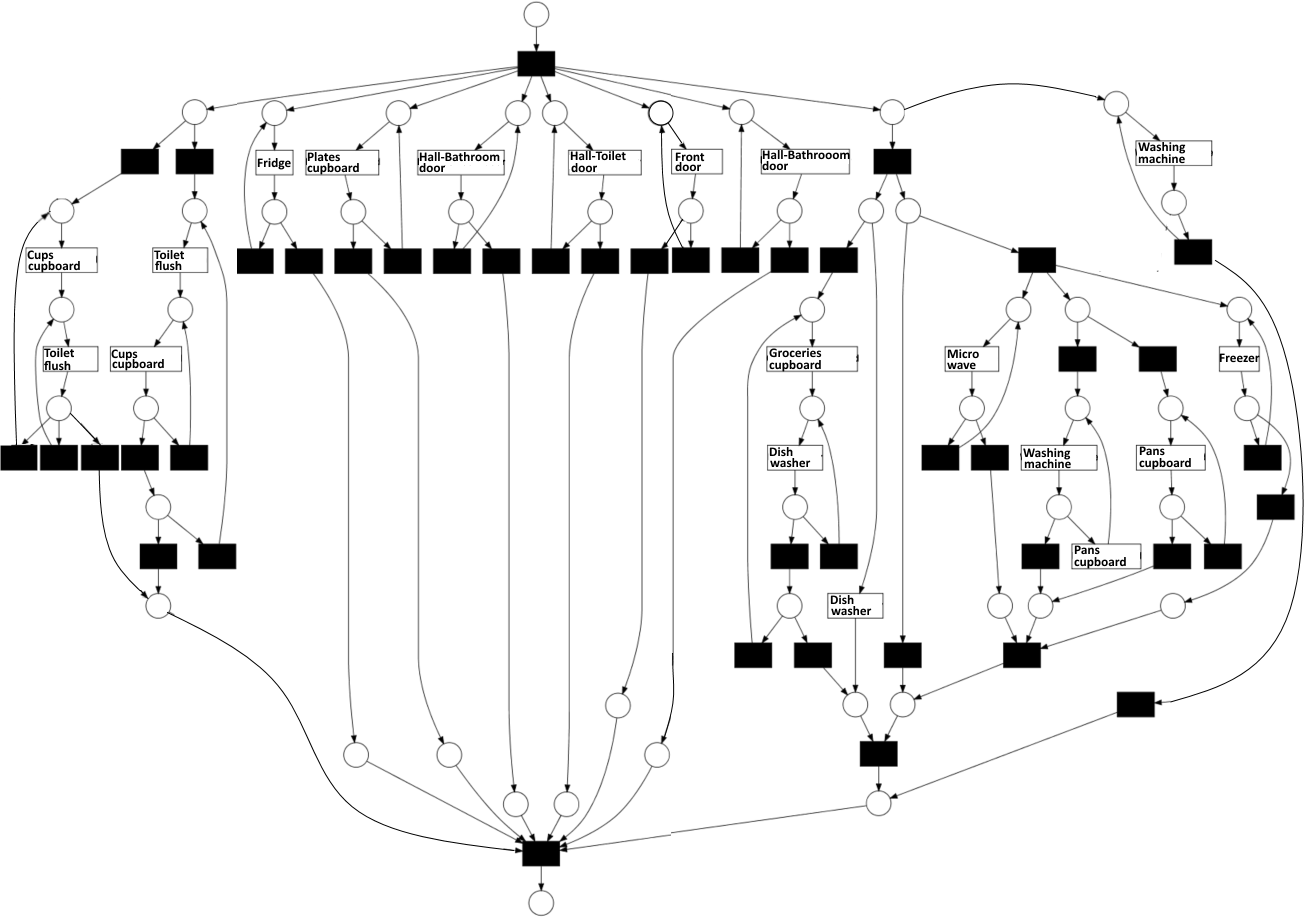}
		\caption{Inductive Miner result on the low-level events from the low-level Van Kasteren event log.}
		\label{fig:kasteren_no_abstraction}
	\end{subfigure}
	\begin{subfigure}{0.8\textwidth}
		\centering
		\includegraphics[width=0.8\textwidth]{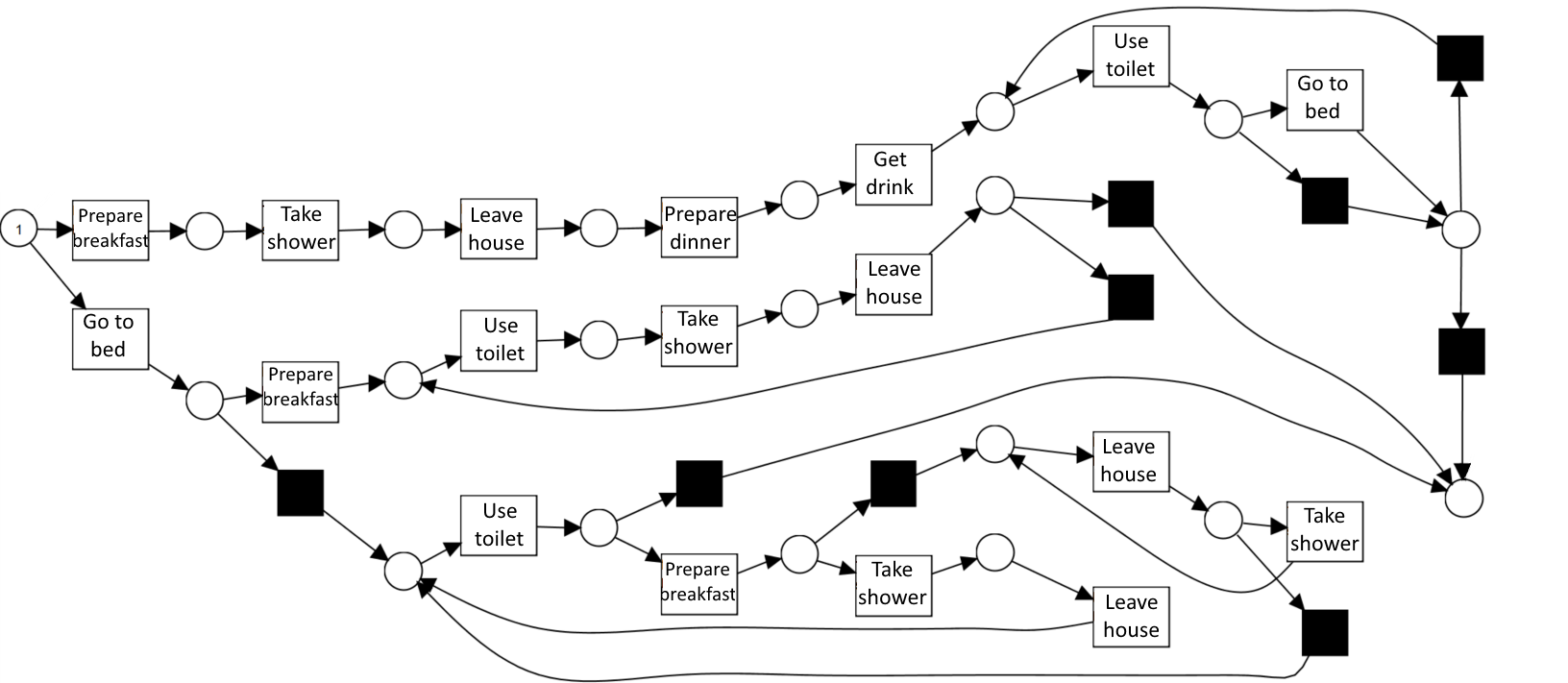}
		\caption{Inductive Miner result on the high-level events discovered from the low-level Van Kasteren event log.}
		\label{fig:kasteren_abstraction}
	\end{subfigure}
	\caption{Comparison of process models discovered from the low-level and high-level van Kasteren event log.}
\end{figure*}

We use the smart home environment log described by Van Kasteren et al. \cite{Kasteren2008} to evaluate our supervised event log abstraction method. The Van Kasteren log consists of multidimensional time series data with all dimensions binary, where each binary dimension represents the state of an in-home sensor. These sensors include motion sensors, open/close sensors, and power sensors (discretized to $0$/$1$ states).

\subsection{Experimental setup}
We transform the multidimensional time series data from sensors into events by regarding each sensor change point as an event. Cases are created by grouping events together that occurred in the same day, with a cut-off point at midnight. High-level labels are provided for the Van Kasteren data set.

The generated event log based on the Van Kasteren data set has the following XES extensions:
\begin{description}[\IEEEsetlabelwidth{Lifecycle}]
	\item[Concept]{The sensor that generated the event.}
	\item[Time]{The time stamp of the sensor change point.}
	\item[Lifecycle]{\emph{Start} when the event represents a sensor value change from $0$ to $1$ and \emph{Complete} when it represents a sensor value change from $1$ to $0$.}
\end{description}

Note that annotations are provided for all traces in the obtained event log. To evaluate how well the supervised event abstraction method generalized to unannotated traces, we artificially use a part of the traces to train the abstraction model and apply them on a test set where we regard the annotations to be non-existent. We evaluate the obtained high-level labels against the ground truth labels. We use a variation on Leave-One-Out-Cross-Validation where we leave out one trace to evaluate how well this mapping generalizes to unseen events and cases.

\subsection{Results}
Figure \ref{fig:kasteren_no_abstraction} shows the result of the Inductive Miner \cite{Leemans2013} for the low-level events in the Van Kasteren data set. The resulting process model starts with many parallel activities that can be executed in any order and contains many unobservable transitions back. This closely resembles the flower model, which allows for any behavior in any arbitrary order. From the process model we can learn that \emph{toilet flush} and \emph{cups cupboard} frequently co-exists. %, as the both are frequently eventually followed by the other activity.
Furthermore, the process model indicates that \emph{groceries cupboard} is often followed by \emph{dishwasher}. There seems to be very little structure on this level of event granularity.

The average Levenshtein similarity between the predicted high-level traces in the Leave-One-Trace-Out-Cross-Validation experimental setup and the ground truth high-level traces is $0.7042$, which shows that the supervised event abstraction method produces traces which are fairly similar to the ground truth.

Figure \ref{fig:kasteren_abstraction} shows the result of the Inductive Miner on the aggregated set of predicted test traces. %This high-level process model is much more structured than the model in Figure \ref{fig:kasteren_no_abstraction} that we discovered on the low-level event log.
Figure \ref{fig:kasteren_abstraction} shows that the process discovered at the high level of granularity is more structured than the process discovered at the original level of granularity (Figure \ref{fig:kasteren_no_abstraction}). In Figure \ref{fig:kasteren_abstraction}, we can see that the main daily routine starts with breakfast, followed by a shower, after which the subject leaves the house to go to work. After work the subject prepares dinner and has a drink. The subject mainstream behavior is to go to the toilet before going to bed, but he can then wake up later to go to the toilet and then continue sleeping. Note that the day can also start with going to bed. This is related to the case cut-off point of a trace at midnight. Days when the subject went to bed after midnight result in a case where going to bed occurs at the start of the trace. On these days, the subject might have breakfast and then perform the activity sequence use toilet, take shower, and leave house, possibly multiple times. Another possibility on days when the subject went to bed after midnight is that he starts by using the toilet, then has breakfast, then has the possibility to leave the house, then takes a shower, after which he always leaves the house. Prepare dinner activity is not performed on these days.

This case study shows that we can find a structured high-level process from a low-level event log where the low-level process is unstructured, using supervised event abstraction and process discovery.

\section{Case Study 2: Artificial Digital Photocopier}
\label{sec:case_2}

\begin{figure*}
	\centering
	\begin{subfigure}{0.8\textwidth}
		\centering
		\includegraphics[width=0.85\textwidth]{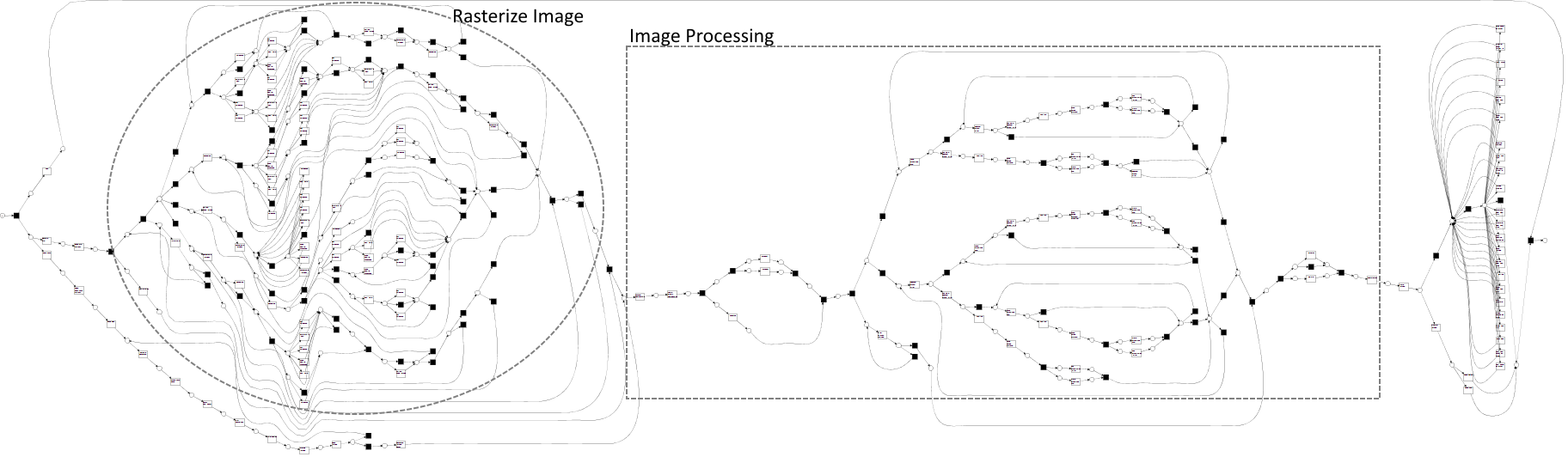}
		\caption{Inductive Miner result on the low-level Artificial Digital Photo Copier event log.}
		\label{adp_no_abstraction}
	\end{subfigure}
	\begin{subfigure}{0.8\textwidth}
		\centering
		\includegraphics[width=0.85\textwidth]{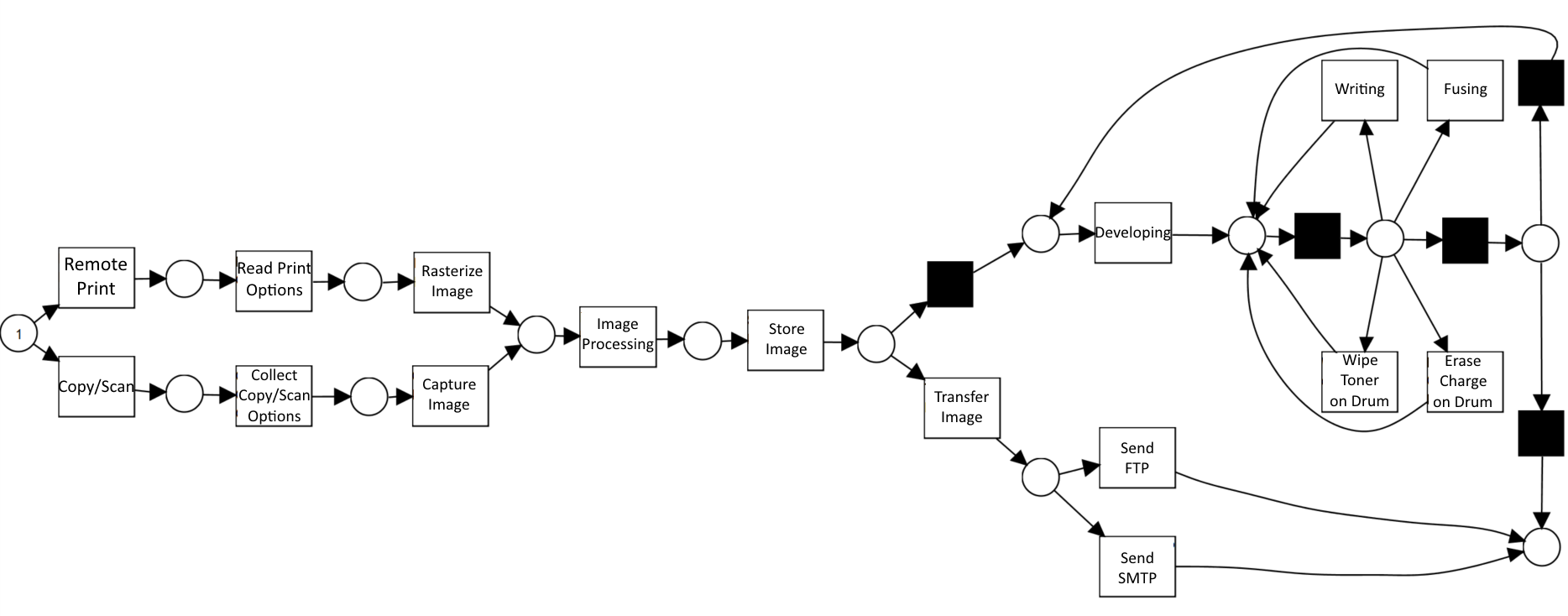}
		\caption{Inductive Miner result on the discovered high-level events on the Artificial Digital Photo Copier event log.}
		\label{adp_abstraction}
	\end{subfigure}
	\caption{Comparison of process models discovered from the low-level and high-level Artificial Digital Photo Copier event log.}
\end{figure*}

Bose et al. \cite{Bose2012,Bose2012b} created a synthetic event log based on a digital photocopier to evaluate his unsupervised methods of event abstraction. In this case study we show that the described supervised event abstraction method can accurately abstract to high-level labels.
\subsection{Experimental setup}
We annotated each low-level event with the correct high-level event using domain knowledge from the actual process model as described by Bose et al. \cite{Bose2012,Bose2012b}. This event log is generated by a hierarchical process, where high-level events \emph{Capture Image}, \emph{Rasterize Image}, \emph{Image Processing} and \emph{Print Image} are defined in terms of a process model. The \emph{Print Image} subprocess amongst others contains the events \emph{Writing}, \emph{Developing} and \emph{Fusing}, which are themselves defined as a subprocess. In this case study we set the task to transform the log such that subprocesses \emph{Capture Image}, \emph{Rasterize Image} and \emph{Image Processing}, \emph{Writing}, \emph{Fusing} and \emph{Developing}. Subprocesses \emph{Writing} and \emph{Developing} both contain the low-level event types \emph{Drum Spin Start} and \emph{Drum Spin Stop}. In this case study we focus in particular on the \emph{Drum Spin Start} and \emph{Drum Spin Stop} events, as they make the abstraction task non-trivial in the sense that no one-to-one mapping from low-level to high-level events exists.

The artificial digital photocopier data set has the concept, time and lifecycle XES extensions. On this event log annotations are available for all traces. On this data set we use a 10-Fold Cross-Validation setting on the traces to evaluate how well the supervised event abstraction method abstracts low-level events to high-level events on unannotated traces, as this data set is larger than the Van Kasteren data set and Leave-One-Out-Cross Validation would take too much time.
\subsection{Results}
The confusion matrix in Table \ref{tab:conf_mat_adp} shows the aggregated results of the mapping of low-level events \emph{Drum Spin Start} and \emph{Drum Spin Stop} to high-level events \emph{Developing} and \emph{Writing}. The results show that the supervised event abstraction method is capable of detecting the many-to-many mappings between the low-level and high-level labels, as it maps these low-level events to the correct high-level event without making errors. The Levenshtein similarity between the aggregated set of test fold high-level traces and the ground truth high-level traces is close to perfect: 0.9667.

\begin{table}
	\centering
	\caption{Confusion matrix for classification of \emph{Drum Spin Start} and \emph{Drum Spin Stop} low-level events into high-level events \emph{Writing} and \emph{Developing}.}
	\label{tab:conf_mat_adp}
	\begin{tabular}{l|ll}
		\multicolumn{1}{l}{} & Developing & Writing \\
		\cline{2-3}
		Developing & 6653 & 0 \\
		Writing & 0 & 917 \\
	\end{tabular}
\end{table}

Figure \ref{adp_no_abstraction} shows the process model obtained with the Inductive Miner on the low-level events in the artificial digital photocopier dataset. The two sections in the process model that are surrounded by dashed lines are examples of high-level events within the low-level process model. Even though the low-level process contains structure, the size of the process model makes it hard to comprehend. Figure \ref{adp_abstraction} shows the process model obtained with the same process discovery algorithm on the aggregated high-level test traces of the 10-fold cross validation setting. This model is in line with the official artificial digital photocopier model specification, with the \emph{Print Image} subprocess unfolded, as provided in \cite{Bose2012,Bose2012b}. In contrast to the event abstraction method described by Bose et al. \cite{Bose2012b} which found the high-level events that match specification, supervised event abstraction is also able to find suitable event labels for the generated high-level events. This allows us to discover human-readable process models on the abstracted events without performing manual labeling, which can be a tedious task and requires domain knowledge.

Instead of event abstraction on the level of the event log, unsupervised abstraction methods that work on the level of a model (e.g. \cite{Vanhatalo2009}) can also be applied to make large complex models more comprehensible. Note that such methods also do not give guidance on how to label resulting transitions in the process model. Furthermore, such methods do not help in cases where the process on a low-level is unstructured, like in the case study as described in Section \ref{sec:case_1}.

This case study shows that supervised event abstraction can help generating a comprehensible high-level process model from a low-level event log, when a low-level process model would be too large to be understandable.

\section{Conclusion}
\label{sec:conclusion}
In this paper we described a method to abstract events in a XES event log that is too low-level, based on supervised learning. The method consists of an approach to generate a feature representation of a XES event, and of a Conditional Random Field based learning step. An implementation of the method described has been made available as a plugin to the ProM 6 process mining toolkit. We introduced an evaluation metric for predicted high-level traces that is closer to process mining than time-window based methods that are often used in the sequence labeling field. Using a real life event log from a smart home domain, we showed that supervised event abstraction can be used to enable process discovery techniques to generate high-level process insights even when process models discovered by process mining techniques on the original low-level events are unstructured. Finally, we showed on a synthetic event log that supervised event abstraction can be used to discover smaller, more comprehensible, high-level process models when the process model discovered on low level events is too large to be interpretable.

% conference papers do not normally have an appendix

% use section* for acknowledgement
%\section*{Acknowledgment}

% trigger a \newpage just before the given reference
% number - used to balance the columns on the last page
% adjust value as needed - may need to be readjusted if
% the document is modified later
%\IEEEtriggeratref{8}
% The "triggered" command can be changed if desired:
%\IEEEtriggercmd{\enlargethispage{-5in}}

% references section

% can use a bibliography generated by BibTeX as a .bbl file
% BibTeX documentation can be easily obtained at:
% http://www.ctan.org/tex-archive/biblio/bibtex/contrib/doc/
% The IEEEtran BibTeX style support page is at:
% http://www.michaelshell.org/tex/ieeetran/bibtex/
%\bibliographystyle{IEEEtran}
% argument is your BibTeX string definitions and bibliography database(s)
%\bibliography{IEEEabrv,../bib/paper}
\bibliographystyle{IEEEtran}
\bibliography{IEEEabrv,bibliography}
% <OR> manually copy in the resultant .bbl file
% set second argument of \begin to the number of references
% (used to reserve space for the reference number labels box)

% that's all folks
\end{document}